\documentclass{article}

\pdfoutput=1 

\usepackage[preprint,nonatbib]{neurips_2022}

\usepackage{mathtools}
\usepackage{xspace}
\usepackage{amssymb}
\usepackage[utf8]{inputenc} 
\usepackage[T1]{fontenc}    
\usepackage{hyperref}       
\usepackage{url}            
\usepackage{booktabs}       
\usepackage{amsfonts}       
\usepackage{nicefrac}       
\usepackage{microtype}      
\usepackage[dvipsnames]{xcolor}         
\usepackage{bm}
\usepackage{tikz}
\usepackage{sistyle}
\SIthousandsep{,}
 \usepackage{arabtex}
\renewenvironment{abstract}%
{%
  \vskip 0.075in%
  \centerline%
  {\large\bf Abstract}%
  \vspace{0.5ex}%
  \begin{quote}%
}
{
  \par%
  \end{quote}%
  \vskip 1ex%
}

\usepackage{algorithm}
\usepackage{multicol}
\usepackage[noend]{algpseudocode}
\algnewcommand\algorithmicforeach{\textbf{for each}}
\algdef{S}[FOR]{ForEach}[1]{\algorithmicforeach\ #1\ \algorithmicdo}
\usepackage{tikz}
\usepackage{todonotes} 
\usetikzlibrary{matrix}
\usetikzlibrary{positioning}
\usepackage{tablefootnote}
\usepackage{booktabs} 
\usepackage{color, colortbl} 
\usepackage{subcaption}
\usepackage{pgfplots}
\usepgfplotslibrary{groupplots}
\usepackage{bbm} 

\usepackage{amsthm}

\newtheorem{theorem}{Theorem}
\newtheorem{lemma}{Lemma}

\usepackage{multirow}

\renewcommand{\l}[0]{k} 

\newcommand{\one}[1]{g} 
\newcommand{\Bias}[0]{\textsc{bias}} 

\newcommand{\x}{\mathbf{X}}

\newcommand{\y}{\mathbf{y}}
\newcommand{\yp}{\tilde{\mathbf{y}}}

\newcommand{\intv}{\mathbb{\intdomain}}

\newcommand{\h}{h}
\newcommand{\hp}{\tilde{h}}

\newcommand{\rr}{\epsilon} 
\newcommand{\intdomain}{\mathbb{IR}}

\newcommand{\matrprod}{\mathbf{z}} 
\newcommand{\matrprodscal}{z}
\newcommand{\matrprodc}{\mathbf{C}} 

\newcommand{\thetaset}{\Theta}

\newcommand{\V}{V}
\newcommand{\yu}{\y^u}
\newcommand{\yl}{\y^l}
\newcommand{\yuscal}{y^u}
\newcommand{\ylscal}{y^l}

    \usetikzlibrary{
        pgfplots.colorbrewer,
    }
    \pgfplotsset{
        cycle list/.define={my marks}{
            every mark/.append style={dotted,fill=\pgfkeysvalueof{/pgfplots/mark list fill}},mark=none\\
            every mark/.append style={solid,fill=\pgfkeysvalueof{/pgfplots/mark list fill}},mark=square*\\
            every mark/.append style={solid,fill=\pgfkeysvalueof{/pgfplots/mark list fill}},mark=none\\
            every mark/.append style={solid,fill=\pgfkeysvalueof{/pgfplots/mark list fill}},mark=diamond*\\
        },
    }

\DeclareMathOperator*{\argmax}{argmax}

\usepackage[capitalize]{cleveref}

\crefname{section}{\S}{section}

\usetikzlibrary{external}
\usepackage[backend=bibtex]{biblatex}
\title{Certifying Data-Bias Robustness in Linear Regression}

\author{%
  Anna P. Meyer, Aws Albarghouthi\thanks{~Author's name in native alphabet: \novocalize\RL{'aws albr.gU_ty}}, and Loris D'Antoni\\
  Department of Computer Sciences\\
  University of Wisconsin--Madison\\
  Madison, WI 53706 \\
  \texttt{\{annameyer}, \texttt{aws}, \texttt{loris\}@cs.wisc.edu} \\
}

\renewcommand{\paragraph}[1]{\textbf{#1.}}
\newtheorem{example}{Example}[section]
\renewcommand{\leq}{\leqslant}
\pgfplotsset{compat=1.17}
\usepackage{fixltx2e}
\begin{document}

\maketitle

\begin{abstract}
  Datasets typically contain inaccuracies due to human error and societal biases, and these inaccuracies can affect the outcomes of models trained on such datasets. We present a technique for certifying whether linear regression models are \emph{pointwise-robust} to label bias in the training dataset, i.e., whether bounded perturbations to the labels of a training dataset result in models that change the prediction of test points.
  We show how to solve this problem exactly for individual test points, and provide an approximate but more scalable method that does not require advance knowledge of the test point.  We extensively evaluate both techniques and find that linear models -- both regression- and classification-based -- often display high levels of bias-robustness. 
  However, we also unearth gaps in bias-robustness, such as high levels of non-robustness for certain bias assumptions on some datasets. Overall, our approach can serve as a guide for when to trust, or question, a model's output.
  \end{abstract}

\section{Introduction}

We see an inherent tension between the growing popularity of machine learning models and the fact that the data that powers these models is not always objective, accurate, or complete \cite{paullada}. Labels of a dataset are particularly vulnerable due to social biases in crowdworkers and other data-labellers, and subjectivity in many labeling tasks \cite{aroyo, hube, sen, tsipras}.  
A key concern, and one we show how to address, is \emph{whether biases or inaccuracies in the data change test-time predictions.} We posit that considering the impact that bias in datasets has on the predictions' \emph{robustness} is key. We illustrate this idea with the following example. Both the example and our approach use linear regression models.

\begin{example}\label{ex:intro1}
Suppose a company $C$ trains a linear regression model to decide what to pay their new hires based on data such as their education and work experience. $C$ wants to compensate its new employees fairly -- in particular, it does not want the salary offers to be tainted by bias in how its current employees (i.e., the training data) are compensated, whether due to the gender pay gap, inconsistencies on how managers apply raises, or other factors. 

Alice was just hired at C and the model predicts that her salary should be \$\num{60000}.
Now, suppose that we could \emph{unbias} the training data, that is, eradicate the gender pay gap, standardize compensation practices across divisions, and remove any other irregularities. If we trained a new model on this unbiased data, would Alice's new salary prediction be the same as (i.e., within $\rr$ of) the old prediction?
If so, then we would say that the company's model is \emph{pointwise-bias-robust} for Alice. 
\end{example}

In this paper, we formally study how biases or inaccuracies in the labels of a training dataset affect the test-time predictions of linear regression models. A na\"ive approach is to enumerate all possible sets of labels, train a model using each, and compare the results. We have two key insights that allow us to solve the problem more efficiently: first, given a fixed test point $\mathbf{x}$, we can take advantage of the linearity of the regression algorithm to exactly compute the minimal label perturbations needed to change the prediction of $\mathbf{x}$ by a given threshold. 
Second, in the absence of a fixed test point, we can extend the same approach and use interval arithmetic to compute an over-approximation of the feasible range of models, allowing for fast (though over-approximate) robustness certification at test time. We discuss how these innovations can help model designers evaluate whether their models' decisions can be trusted, i.e., whether their models are ready to be deployed.

\paragraph{Contributions}
Our main contributions are:
(1) Defining the bias-robustness problem for regression datasets with label bias (\cref{sec:3}).  
(2) A technique to find the smallest training-data label perturbation that changes the prediction of a fixed test point, thereby certifying or disproving pointwise robustness (\cref{sec:exact}). 
(3) A faster, over-approximate certification technique that can certify pointwise bias robustness by bounding the range of linear models obtainable with a fixed amount of label perturbation (\cref{sec:approx}).
(4) Evaluating our results on a variety of datasets and with a variety of forms of bias, and discuss what dataset properties impact label-bias robustness (\cref{sec:experiments}).

\section{Related work}

\paragraph{Uncertainty in machine learning}
Uncertainty in machine learning has been studied in conjunction with fair and trustworthy AI because in high-impact domains it is crucial to understand when an algorithm might fail or have disparate impacts on different subgroups~\cite{ali-uncertainty, bhatt,  tomsett-calibration}. In this area, previous work has shown that many models can yield comparable accuracy (but different predictions) on the same training set \cite{amour-underspecification, fisher-rashomon, predictive-multiplicity}. We conduct a similar  analysis in a parallel direction by considering the range of models that can be learned over similar datasets.

\paragraph{Poisoning attacks and defenses}
Poisoning attacks, where an attacker can modify a small portion of the training dataset, can be effective at reducing the accuracy of models at test-time~\cite{biggio2012poisoning, shafahi2018poison, zhang-label-flipping, xiao-label-flipping}. Various defenses counteract these attacks~\cite{jia-bagging, paudice-label-sanitation,rosenfeld-randomized, steinhardt-poisoning,zhang-debugging}, including ones that focus on attacking and defending linear regression models~\cite{jagielski-poisoning, muller-poisoning}. Some of these works have a broad definition of bias that includes label perturbations, data deletion, and data poisoning. We only consider label perturbations, making our threat model similar to a handful of others \cite{paudice-label-sanitation, rosenfeld-randomized, xiao-label-flipping}; however, we allow a more flexible threat model by allowing the user to target the label poisoning to certain subgroups.

Our work differs from existing poisoning literature in two important ways: first, we certify \emph{pointwise} robustness \emph{deterministically}: that is, we prove, with 100\% certainty, whether the prediction of a particular test point is vulnerable to bias (i.e., poisoning) in the training data. Second, we do not modify the training procedure, as in~\cite{jia-bagging, steinhardt-poisoning, rosenfeld-randomized}, nor do we make assumptions that we have access to (parts of) the clean data~\cite{steinhardt-poisoning, zhang-debugging}. Meyer et al. have a similar problem definition to ours: they certify robustness by overapproximating the set of models that one can learn on variants of the dataset, and although their approach applies to complex forms of bias, it is limited to decision trees~\cite{meyer}.

\paragraph{Robustness} Distributional robustness studies how to find models that perform well across a family of distributions~\cite{ben-tal-distribution, namkoong-distribution, shafieezadeh-distribution}. Our problem definition can be framed as working over a family of distributions (where the distributions differ only in the labels). Robust statistics shows how algorithms can be adapted to account for outliers or other errors in the data~\cite{diakonikolas2019recent, diakonikolas_robustness}. Our work does not require that the data follows any particular distribution, and thus we can certify a broader range of datasets. Others have shown that some algorithms are inherently, or modifiable to be, robust to label noise~\cite{natarajan, patrini, rolnick}. However, these works -- along with the robust statistics and distributional robustness literature -- provide statistical global robustness guarantees, rather than the provable exact robustness guarantees that we make. That is, they are interested in finding one good classifier, rather than understanding the range of classifiers that could be obtained even in extreme cases.

\paragraph{Algorithmic stability} Algorithmic stability measures how sensitive algorithms are to small perturbations in the training data~\cite{breiman-stability, devroye-stability}. Our work differs from this literature in a couple ways: first, rather than the commonly used leave-one-out perturbation model~\cite{black-leaveoneout, kearns-stability}, we allow for a fixed number of small label perturbations; second, as discussed above, our focus is certifying point-wise robustness rather than ensuring comparable overall behavior.

\section{Label-Bias Robustness }\label{sec:3}

First, we formally define label bias and what it means for a learning algorithm to be bias-robust.


\paragraph{Label-bias function} 
We define a dataset as a matrix of features $\x \in \mathbb{R}^{n \times m}$
and a corresponding vector of labels $\y \in \mathbb{R}^n$.
The \emph{label-bias function}, or bias function for short, $\Bias: \mathbb{R}^n\rightarrow 2^{\mathbb{R}^n}$, is a function that takes in a label vector $\y$ and returns set of 
label vectors such that $\y \in \Bias(\y)$. 
Intuitively, $\Bias(\y)$ returns a set of all labels vectors that are similar to $\y$ (we will define what ``similar'' means in \cref{sec:models_of_bias}). Thus, if $\y$ is biased by a small amount, its ``true,'' unbiased counterpart will be in $\Bias(\y)$. We will refer to $\Bias(\y)$ as the label-bias set of $\y$, or bias set for short.

\paragraph{Pointwise label-bias robustness} 
Assume we have a learning algorithm that, given a training
dataset, returns a model $\h$ from some hypothesis class. 
Fix a dataset $(\x,\y)$ and bias function $\Bias$. 
Let $\h$ be the model learned from $(\x,\y)$.
Given a test point $\mathbf{x}\in\mathbb{R}^{m}$ and a \emph{robustness radius} $\rr$, we say that our learning algorithm is \emph{$(\rr,\Bias)$-robust} on $\mathbf{x}$
iff for all $\yp \in \Bias(\y)$, if $\hp$ is the model learned on $(\x,\yp)$, then
\begin{equation}\label{eq:rob}
\hp(\mathbf{x}) \in [\h(\mathbf{x}) - \rr, \h(\mathbf{x}) + \rr]
\end{equation}
In other words, even if there were label bias in the dataset,
the prediction on $\mathbf{x}$ would not deviate by more than $\rr$.

\subsection{Models of Bias}\label{sec:models_of_bias}
We now demonstrate how we define the bias function $\Bias$ to mimic real-world scenarios.

\paragraph{Vectors of intervals}
We will use interval arithmetic to define $\Bias$.
The set of intervals is defined as 
$\intv = \{[l,u]\mid l,u\in\mathbb{R},\ l\leq u\}$. 
Given an interval $\V=[l,u]$, we define $\V^l=l$ and $\V^u=u$.
If $\Delta \in \intv^n$ is a vector of intervals,
we use $\delta_i^l$ and $\delta_i^u$ to denote the lower and upper bounds
of the $i$th interval in $\Delta$.

\paragraph{Defining the bias set} 
We assume that up to $\l$ of the labels in $\y$ are incorrect, that is, perturbed within some range. We will use a vector of intervals $\Delta$ to specify how much each
label can be changed---we call this vector the \emph{perturbation vector}.

Given a dataset $(\x,\y)$, a maximum number $\l$ of biased labels, 
and a perturbation vector $\Delta$, 
the set $\Bias_{\l, \Delta}(\y)$ is the set of all label vectors that we can create by modifying up to $\l$ labels in $\y$ according to $\Delta$,
i.e., 
$$ \Bias_{\l, \Delta}(\y) = \{ \yp \mid \tilde{y}_i \in [y_i + \delta_i^l , y_i + \delta_i^u] \textrm{ for all } i \textrm{ and } ||\yp-\y||_0 \leq \l\} $$

Since we require that $\y\in\Bias(\y)$, it is necessary that $0\in\Delta$.

\begin{example}\label{ex:toy1_con}
Let $\Delta= ([-1,1],[-1,1])^\top$. 
Given $\y=(3,4)^\top$ and a maximum number of biased labels $\l=1$, we have 
$$ \Bias_{\l,\Delta}(\y) = \left\{ \y \right\} \cup \left\{\begin{pmatrix} a \\ 4  \end{pmatrix}\mid a\in[2,4]\right\} \cup \left\{ \begin{pmatrix} 3 \\ b  \end{pmatrix} \mid b\in[3,5]\right\}.$$
In other words, at most one label in $\y$ can be changed by at most 1.
\end{example}

\begin{example}
We continue the premise from \cref{ex:intro1}.
If we know, or suspect, that C has a gender wage gap (i.e., women are systematically paid less than men), we can model $\Delta$ as follows: if the gender of data point $\mathbf{x}_i$ is male, then $\delta_i=[0,0]$; if the gender of $\mathbf{x}_i$ is female, then $\delta_i=[0,2000]$, under the assumption that each woman employee is potentially underpaid by up to \$\num{2000}. 
\end{example}

\paragraph{Binary classification} In a classification setting with $\{0,1\}$ labels, the default value for $\Delta$ is $\delta_i = [-1,0]$ when $y_i=1$, and $\delta_i=[0,1]$ when $y_i=0$, modeling a label flip. If we wish to further limit the bias, these intervals may be changed to $[0,0]$ for some indices. 
E.g., if we believe that there was historical bias against women leading to incorrect negative labels, then we can use $\delta_i=[0,1]$ for women with a negative label (i.e., when $y_i = 0$),
and $\delta_i=[0,0]$ for the rest of the dataset.


\section{Exact Bias-Robustness Certification for Linear Regression}\label{sec:exact}

In this section, we present a technique for proving bias robustness (\cref{eq:rob})
for a given input.
Our approach is an exact certification technique, that is, it can prove or disprove $(\rr,\Bias)$-robustness.
In the latter case, it will produce a perturbation $\yp \in \Bias_{\l,\Delta}(\y)$
that changes the prediction by more than $\rr$, as we illustrate in the following example.

\begin{example}\label{ex:exact1}
We continue the premise from \cref{ex:intro1}. \Cref{fig:intro_ex}a shows salary data and the resulting regression model (all features are projected onto the $x$-axis). \Cref{fig:intro_ex}b shows the same dataset, with one salary (in red) perturbed by $+\$\num{10000}$. Note that the new model is slightly different from the original. \Cref{fig:intro_ex}c shows a data point, $\mathbf{x}$, that -- given a robustness radius of \$1000 --  is not robust since the original model predicts $\mathbf{x}$'s salary as \$\num{85720} while the new one predicts \$\num{90956}.
\end{example}


\paragraph{Linear regression}
We use least-squares regression, i.e., we want to solve the optimization problem 
$$ \min_{\theta\in\mathbb{R}^m} (\y - \x\theta) $$
Least-squares regression admits a closed-form solution, $\theta = (\x^\top \x)^{-1} \x^\top \y$, for which we will prove bias-robustness.
We work with the closed-form solution, instead of a gradient-based one, as it is deterministic and holistically considers the whole dataset, allowing us to exactly certify bias robustness by exploiting linearity (Rosenfeld et al. make an analogous observation~\cite{rosenfeld-randomized}, and on medium-sized datasets and modern machines, computing the closed-form solution is efficient).

Given an input $\mathbf{x}$, we predict its label $\hat{y}$ as $\theta^\top\mathbf{x}$.
We can expand and rearrange $\theta^\top \mathbf{x}$ as follows:
\begin{align*}
    \theta^\top \mathbf{x} &= ((\x^\top \x)^{-1} \x^\top \y )^\top \mathbf{x} \\
                     & = \underbrace{(\mathbf{x}^\top (\x^\top \x)^{-1}\x^\top)}_{\matrprod}\y
\end{align*}
We will use $\matrprod$ to denote $\mathbf{x}^\top (\x^\top \x)^{-1}\x^\top$.

\paragraph{Certifying robustness for regression tasks} Suppose we have a dataset $(\x,\y)$, a corresponding bias model $\Bias_{\l,\Delta}(\y)$,
and a test point $\mathbf{x}$.
We want to prove that the prediction on $\mathbf{x}$ is $(\rr,\Bias_{\l,\Delta})$-robust,
as per \cref{eq:rob}.
Informally, our goal is to find how the prediction $\hat{y}=\theta^\top \mathbf{x}$ varies when different $\yp\in\Bias_{\l,\Delta}(\y)$ are used to find $\theta$. 
Specifically, we aim to find 
\begin{equation}\label{eq:max}
    \max_{\yp\in\Bias_{\l,\Delta}(\y)} |\matrprod \yp - \matrprod\y| 
\end{equation}

\begin{lemma}\label{lemma:1}
Given $\yp$ as per \cref{eq:max}, 
$\mathbf{x}$ is $(\rr,\Bias_{\l,\Delta})$-robust iff $\matrprod\yp\in[\hat{y}-\rr, \hat{y}+\rr]$.
\end{lemma}

\paragraph{Algorithmic technique}
The na\"ive approach to solve this maximization objective is to enumerate all $\yp\in\Bias_{\l,\Delta}(\y)$, train a model on each, and compare the resulting $\hat{y}$'s. Of course, this will not be computationally feasible since the set $\Bias_{\l,\Delta}(\y)$ is infinite.
Instead, we exploit the linearity of the inner product $\matrprod \y$ and iteratively find a label $y_i$ that when changed will result in the largest perturbation to the prediction of $\mathbf{x}$. 

We will discuss how to find the range of values $V$ that $\matrprod\yp$ can take on, i.e., 
$$ 
    \V = \left[ \min_{\yp\in\Bias_{\l,\Delta}(\y)} \matrprod\yp, \max_{\yp\in\Bias_{\l,\Delta}(\y)} \matrprod\yp \right]
$$
To check robustness, we check whether $\V\subseteq[\matrprod\y - \rr, \matrprod\y+\rr]$. Formally:
Formally,

\begin{lemma}\label{lemma:2}
$\V\subseteq[\matrprod\y - \rr, \matrprod\y+\rr]$ iff $\mathbf{x}$ is $(\rr, \Bias_{\l,\Delta})$-robust
\end{lemma}

We discuss how to find the upper bound of $V$ here---see \cref{alg:exact}. 
Finding the lower bound is analogous and is presented in the appendix. 

Intuitively, the labels in $\y$ with the highest influence on the prediction of 
$\mathbf{x}$ are those corresponding to elements of $\matrprod$ with the largest magnitude.
However, we must also take into account $\Delta$: even if $\matrprodscal_i$ is very large, perturbing $y_i$ will only affect the outcome if $\delta_i$ allows for a non-zero perturbation.

We define the \emph{positive potential impact} $\rho_i^+$ of  $\matrprodscal_i$ as the maximal positive change that perturbing $y_i$ can have on $\matrprod\y$.
After computing $\rho^+$, we find the $\l$ largest elements of $\rho^+$ by absolute value. The indices of these entries correspond to the indices of the labels of $\y$ we will change to maximally increase $\matrprod\y$, i.e., to find the upper bound of $\V$.

\begin{example}\label{ex:ppi}
Let $\y=(3,4)^\top$ and $\matrprod=(-1,2)$. Suppose $\Delta = ([-1,1],[-1,1])^\top$. Since $\matrprodscal_1< 0$, we can increase $\matrprod\y$ by decreasing $y_1$, that is, by adding $\delta_1^l=-1$ to it. Thus $\rho^+_1 = \matrprodscal_1 \delta_1^l = 1$. Since $\matrprodscal_2\geq 0$, we can increase $\matrprod\y$ by increasing $y_2$. Therefore, $\rho^+_2 = \matrprodscal_2\delta_2^l = 2$. 

\end{example}


\input{graphs/intro_ex} 

\begin{algorithm}[!t]
    \caption{Find the upper bound of $\V = [ \min_{\yp\in\Bias_{\l,\Delta}(\y)} \matrprod\yp, \max_{\yp\in\Bias_{\l,\Delta}(\y)} \matrprod\yp]$
    }
    \begin{algorithmic}
        \Require $\matrprod\in\mathbb{R}^n, \y\in\mathbb{R}^n, \Delta\in\intdomain^n$ with $0\in\Delta$, $\l\geq 0$
        \State $\yu\gets\y$
        \If{$\matrprodscal_i\geq 0$}
             $\rho^+_i \gets \matrprodscal_i\delta_i^u\ $
        \textbf{else}
             ~$\rho^+_i \gets \matrprodscal_i\delta_i^l\ $
        \EndIf
        \State Let $\rho^+_{i_1},\ldots,\rho^+_{i_\l}$ be the $\l$ largest elements of $\rho^+$ by absolute value
        \ForEach{$\rho^+_{i_j}$}
            \If{$\matrprodscal_{i_j}\geq 0$}
                 ${\yuscal_{i_j}} \gets {\yuscal_{i_j}} + \delta_{i_j}^u$
            \textbf{else}~
                 ${\yuscal_{i_j}}\gets {\yuscal_{i_j}} + \delta_{i_j}^l$
            \EndIf
        \EndFor
        \State $\V^u \gets \matrprod\yu$
    \end{algorithmic}\label{alg:exact}
\end{algorithm}

\begin{example}\label{ex:sec4}
We continue \cref{ex:ppi}. Suppose $\l=1$. We initially set 
$\yu=\y$. Since $\rho^+=(1,2)$, we initially perturb $\yuscal_2$. 
Since ${\matrprodscal}_2\geq 0$, we add $\delta_2^u$ to $\yuscal_2$, that is, we set ${\yuscal_2} = 4+1=5$. 
Thus, with $\yu=(3,5)$,
we calculate $\V^u = \matrprod\yu = 7$.
Similar logic will find $V^l=3$.

Suppose $\epsilon=3$. We need to check whether $\V\subseteq[\matrprod\y-\rr, \matrprod\y+\rr] = [2,8]$. Since $\V=[3,7]$, this condition is satisfied and the input is $\Bias_{\l,\Delta}$-robust.
\end{example}

\begin{theorem}\label{thm:exact}
\cref{alg:exact} computes the upper bound of $V$.
\end{theorem}

\begin{example}
\Cref{fig:intro_ex}c happens to perturb the data point that causes the maximal positive prediction for $\mathbf{x}$, i.e., $V^u=\$\num{90956}$. 
\end{example}

\paragraph{Alternate goal: finding the minimum $\l$} We may be interested in finding the smallest $\l$ such that $\mathbf{x}$ is not robust to $\l$ label perturbations, i.e., the threshold at which $\mathbf{x}$ stops being robust. To find this threshold, we iteratively call \cref{alg:exact} with $\l=1$, while keeping track of $\yu$ and $\rho$ to ensure that we do not perturb the same label multiple times. We check robustness after each iteration -- as soon as the robustness check fails, we end our search. 
If we get to a point where we have perturbed all indices of $\y$, we stop and conclude that no number of label perturbations will cause non-robustness for $\mathbf{x}$.

The next theorem discusses the relationship between $\Delta$ and $\epsilon$. Specifically, we find that the ratio between $\rr$ and $\Delta$ uniquely determines robustness.
We make use of this fact in \cref{sec:experiments}.

\begin{theorem}\label{thm:ratio}
    $\mathbf{x}$ is $(\rr, \Bias_{\l,\Delta})-$robust iff $\mathbf{x}$ is $(c\rr, \Bias_{\l, c\Delta})$-robust for all $c>0$. 
\end{theorem}

\paragraph{Classification}
Our method is also exact in the binary classification setting (details in appendix).
\section{An Approximate Certification Method}\label{sec:approx}

In \cref{sec:exact}, we described a decision procedure for certifying bias-robustness.
For every input $\mathbf{x}$ for which we want to certify robustness, the procedure effectively relearns the worst-case linear regression model for $\mathbf{x}$.
In practice, we may want to certify robustness of a large number of samples, e.g., a whole test dataset, or we may need to perform online certification.
We would like to perform this certification without having to solve linear regression for every input. The following example provides intuition for our approach.

\begin{example} \label{ex:approx_running}
We continue the scenario from \cref{ex:exact1}. 
\Cref{fig:intro_ex}d shows the entire range of linear models obtainable by changing any one data point by $\pm$\$\num{10000}. Given a test point $\mathbf{x}$, the range of predictions is the range of the shaded region at $\mathbf{x}$. 

\end{example}

We formalize capturing all regression models we may obtain as follows:
$$\thetaset = \{\theta\mid \theta = (\x^\top \x)^{-1} \x^\top \yp \textrm{ for some } \yp\in\Bias_{\l,\Delta}(\y)\}$$
To certify robustness of $\mathbf{x}$, we then simply check that every $\tilde{\theta} \in \thetaset$ satisfies $\tilde{\theta}^\top \mathbf{x} \in [\theta^\top \mathbf{x}-\rr, \theta^\top \mathbf{x}+\rr]$.

For ease of notation, let $\matrprodc=(\x^\top\x)^{-1}\x^\top$.
Note that $\matrprodc\in\mathbb{R}^{m\times n}$, while $\matrprod\in\mathbb{R}^{1\times n}$.

\paragraph{Challenges}
The set of weights $\thetaset$ is an infinite, non-convex set (see \cref{ex:approx1}). 
Our goal is to represent $\thetaset$ efficiently so that we can simultaneously apply
all weights $\theta \in \thetaset$ to a point $\mathbf{x}$.
Our key observation is that we can easily compute a hyperrectangular approximation of $\thetaset$.
In other words, we want to compute a set $\thetaset^a$ such that $\thetaset \subseteq \thetaset^a$.
Note that the set $\thetaset^a$ is an interval vector in $\intv^n$, since interval vectors represent hyperrectangles in Euclidean space.

This approximation approach results in an \emph{incomplete} procedure that abstains when it cannot prove robustness of an input.

\paragraph{Approximation approach}
We will iteratively compute component of the vector $\thetaset^a$ 
by finding each coordinate $i$ as the following interval, where $\mathbf{c}_i$ are the column vectors of $\matrprodc$:
$$ 
\thetaset^a_i = \left[\min_{\y'\in\Bias_{\l,\Delta}(\y)} \mathbf{c}_i \y', \max_{\y'\in\Bias_{\l,\Delta}(\y)} \mathbf{c}_i \y'\right]
$$
To find $\min_{\y'\in\Bias_{\l,\Delta}(\y)} \mathbf{c}_i\y'$, we use the same process as in \cref{sec:exact}.
Specifically, we use \cref{alg:exact} to compute the lower and upper bounds of each $\thetaset^a_i$.
Returning to \cref{ex:approx_running}, the dotted red lines in \cref{fig:intro_ex}d show $\thetaset^a$. Notice that this region overapproximates the exact set of obtainable models, which is shown in solid green.

The following theorem states that the interval matrix
$\thetaset^a_i$ 
is a tight overapproximation of the set $\thetaset$.

\begin{theorem}\label{thm:approx}
 (i) $\thetaset \subseteq \thetaset^a$.
(ii) There does not exist a hyperrectangle ${{\thetaset}^a}' \supseteq \thetaset$
where ${{\thetaset}^a}' \subset {\thetaset}^a$.
\end{theorem}

\paragraph{Evaluating robustness}
Given $\thetaset^a$, as described above, an input $\mathbf{x}$ is $(\rr,\Bias_{\l,\Delta})$-robust if
\begin{align}\label{eq:approx}
    (\thetaset^a)^\top \mathbf{x} \subseteq [\theta^\top \mathbf{x} - \rr, \theta^\top \mathbf{x} + \rr]
\end{align} 
Note that $(\thetaset^a)^\top \mathbf{x}$ is computed using standard interval arithmetic,
e.g., $[l,u] + [l',u'] = [l+l' , u+u']$.
Also note that the above is a one-sided check: we can only say that $\mathbf{x}$ is robust, but because $\thetaset^a$ is an overapproximation, if \cref{eq:approx} does not hold,  we cannot conclusively say that $\mathbf{x}$ is not robust.



\section{Experimental Evaluation}
\label{sec:experiments}

We implement both the exact and approximate pointwise certification procedures in Python. 
To speed up the evaluation, we use a high-throughput computing cluster. (We request 8GB memory and 8GB disk, but all experiments are feasible to run on a standard laptop.)
We find that both techniques are valuable, as  the approximate version -- though it loses precision -- is much faster when certifying a large number of data points. Neither approach has a baseline with which to compare, as ours is the first work to address the bias-robustness problem in linear regression using \textit{exact certification}. 

\paragraph{Description of datasets} We evaluate our approaches for both classification and regression tasks. We divide each dataset into train (80\%), test (10\%), and validation (10\%) sets, except when a standard train/test split is provided (i.e., for MNIST 1/7). The classification datasets are the Income task from the FolkTables project (training $n=\num{26153}$ -- we limited the results to one state to have a manageable dataset size) \cite{folktables, census-data}, the COMPAS dataset ($n=\num{4936}$) \cite{compas}, and MNIST 1/7 ($n=\num{13007}$) \cite{mnist}. We also use the Income dataset for regression (``Income-Reg") by predicting the exact salary. 

We also generate a variety of synthetic datasets to explore what datasets properties impact bias-robustness. Details about the synthetic data experiments are available in the appendix. 

\paragraph{Accuracy-robustness tradeoff}
There is a trade-off between accuracy and robustness that is controlled by the regularization parameter $\lambda$ in the least-squares formula $\theta = (\x^\top \x- \lambda I)^{-1}\x^\top\y$.  Larger values of $\lambda$ improve robustness at the expense of accuracy. \Cref{fig:tolerance} illustrates this trade-off. All results below, unless otherwise stated, use a value of $\lambda$ that maximizes accuracy.  

\input{graphs/tols}

\paragraph{Experiment goals} Our core objective is to see how robust linear classifiers are to bias in the training data. We measure this aspect with the \emph{certifiable robustness rate}, that is, the fraction of test points in a dataset that our algorithm can certify as robust. In \cref{sec:exactresults}, we evaluate the exact technique from \cref{sec:exact}, and then in \cref{sec:approxresults}, we compare with the approximate approach from \cref{sec:approx}. Finally, in \cref{sec:varying}, we show how different dataset and bias function properties impact overall bias-robustness. 

\subsection{Exact data-bias robustness}\label{sec:exactresults}
\Cref{tab:all_data} shows the fraction of test points that are bias-robust for classification datasets at various levels of label bias. For each dataset, the robustness rates are relatively high ($>80\%$) when fewer than 0.2\% of the labels are perturbed, and stay above 50\% for 1\% label bias on COMPAS and MNIST 1/7.

\begin{table}[t]
\small
\centering
\caption{Robustness rates (\% certifiably robust) for different numbers of label-flips for various classification datasets given different label perturbation amounts.}
\label{tab:all_data}
\begin{tabular}{l|rrrrrrrrrrr}\toprule
\multirow{2}{*}{Dataset} & \multicolumn{11}{c}{Bias level $\l$ as a percentage of training dataset size}  \\
                         & 0.1  & 0.25  & 0.5  & 0.75  & 1.0  & 1.5  & 2.0  & 3.0  & 4.0 & 5.0 & 6.0 \\\midrule
MNIST 1/7                & 98.3 & 96.3 & 93.1 & 88.3 & 84.4 & 73.1 & 60.8 & 38.8 & 23.3 & 13.1 & 7.0 \\
Income                   & 91.1 & 81.4 & 67.8 & 58.4 & 50.7 & 37.2 & 23.3 & 12.1  & 4.8 & 1.7 & 0.7  \\
COMPAS                   & 96.2 & 91.8 & 85.2 & 78.6 & 72.8 & 57.0 & 46.6 & 31.9 & 17.9 & 6.7 & 3.5 \\

\bottomrule   
\end{tabular}
\end{table}

Despite globally high robustness rates, we must also consider the non-robust data points. In particular, we want to emphasize that for Income and COMPAS, each non-robust point represents an individual whose decision under the model is vulnerable to attack by changing a small number of labels. Some data points will surely fall into this category -- if not, that would mean the model was independent from the training data, which is not our goal! However, if a data point is not robust to a small number of label flips, perhaps the model should not be deployed on that point: instead, the example could be evaluated by a human or auxiliary model. For some datasets, a nontrivial minority of data points are not robust at low bias levels, e.g., at just 0.25\% label bias, around 18.6\% of Income test points are not robust. This low robustness rate calls into question the advisability of using a linear classifier on Income unless one is very confident that label bias is low.

\paragraph{Regression dataset results} \Cref{tab:reg_data} presents results on Income-Reg for the fixed robustness radius $\rr=\$\num{2000}$, which we chose as a challenging, but reasonable,  definition for two incomes being ``close". We also empirically validated that the ratio between $\Delta$ and $\epsilon$ uniquely determines robustness for a fixed bias function (\cref{thm:ratio}). Notably, for small $\Delta$ to $\rr$ ratios, we can certify many test points as robust, even when the number of perturbed training points is relatively large (up to $10\%$). 

\begin{table}[t]
\small
\centering
\caption{Certifiable robustness rates for Income-Reg given various $\Delta$ and $\l$ values. $\epsilon=\num{2000}$ in all experiments. Note that when $\Delta=a$ means $\Delta=[-a,a]^n$. Column 2 gives the ratio between the maximum label perturbation ($\Delta$) and the robustness radius ($\rr$), which uniquely determines robustness.
}
\label{tab:reg_data}
\begin{tabular}{rc|rrrrrrrrrr}\toprule
\multirow{2}{*}{$\Delta$}  & \multirow{2}{*}{Ratio $\frac{\Delta}{\epsilon}$} & \multicolumn{10}{c}{Bias level $\l$ as a percentage of training dataset size} \\ 
    &  & 1.0   & 2.0   & 3.0   & 4.0   & 5.0  & 6.0  & 7.0  & 8.0  & 9.0  & 10.0 \\ \midrule
\num{1000} & 0.5  & 100.0 & 100.0 & 100.0 & 100.0 & 99.8 & 99.4 & 99.1 & 98.6 & 98.0 & 97.1 \\
\num{2000} & 1    & 100.0 & 96.6  & 91.1  & 85.4  & 84.1 & 76.4 & 73.3 & 64.0 & 49.9 & 35.2 \\
\num{4000} & 2    & 91.2  & 84.2  & 69.2  & 35.7  & 14.2 & 2.0  & 0    & 0    & 0    & 0    \\
\num{6000}  & 3    & 85.9  & 61.1  & 15.4  & 0     & 0    & 0    & 0    & 0    & 0    & 0    \\
\num{8000}  & 4    & 80.1  & 20.0  & 0     & 0     & 0    & 0    & 0    & 0    & 0    & 0    \\
\num{10000}  & 5   & 69.6  & 3.6   & 0     & 0     & 0    & 0    & 0    & 0    & 0    & 0    \\ \bottomrule   
\end{tabular}
\end{table}


\subsection{Approximate data-bias robustness}\label{sec:approxresults}
As expected, the approximate approach is less precise than the exact one. The loss in precision depends highly on the dataset and the amount of label bias. \cref{fig:exactapproxcompare} shows that the precision gap is large for MNIST 1/7, relatively small for COMPAS and medium for Income. Unlike in the previous section, we use an accuracy/robustness tradeoff that sacrifices 2\% accuracy to gain robustness.

\input{graphs/exactapproxcompare}

We also measured the time complexity of the exact and approximate approaches.  To certify \num{1000} test points of Income, it takes 44.9 seconds for the exact approach and 6.3 seconds for the approximate approach. For \num{10000} test points, it takes 442.5 seconds and 22.0 seconds, respectively. I.e., the exact approach scales linearly with the number of test points, but the approximate approach stays within a single order of magnitude. See appendix for more details and discussion.

\subsection{Varying dataset properties}\label{sec:varying}
We conduct experiments to probe how the size of the demographic subgroup impacts bias robustness.


\paragraph{Balancing demographic groups}
We discuss trends in robustness of different racial groups (limited to White and Black since these are the largest racial groups in the datasets). We discuss the results from Income below and the results from COMPAS in the appendix.

The original Income dataset we used only includes the U.S. state of Wisconsin, which has a majority-White population. Thus, 92.7\% of the dataset entries have race=White, and just 2.6\% race=Black. We downloaded  other states' data, summarized in the appendix. None of these states have racially-balanced data, however, three of them (Georgia, Louisiana, and Maryland) have more balanced data, and another (Oregon) is similar to Wisconsin, but with a lower share of White people.

\Cref{fig:statesdemo} shows that more race-balanced states have minimal racial robustness gaps. However, the states where Black people were severely underrepresented in the data have a large bias-robustness gap between racial groups. We hypothesize that better demographic representation yields higher robustness, but future work to establish this connection formally is  needed.

\input{graphs/income_states}

\paragraph{Altering $\Delta$ for unequal demographic groups} 
In the United States, structural inequity leads to higher salaries for White people. Our concept of bias functions lets us model this in $\Delta$. We define \emph{targeting} on $\phi$ as limiting label perturbations to samples $\mathbf{x}$ that satisfy $\phi(\mathbf{x})$. \Cref{fig:statestargeting} shows that for Wisconsin, targeting on race=Black (i.e., $\delta_i=0$ when race is not Black) greatly increases certifiable-robustness rates for White people, but barely changes them for Black people. However, in Georgia, targeting on race=Black significantly increases robustness rates for both groups. For both states, targeting on race=White is less impactful.
 
\input{graphs/income_states_targeted} 

There are two takeaways: first, as is common sense, if a demographic group is smaller, there is more uncertainty surrounding that group. E.g., targeting on race=Black for Wisconsin barely changes the robustness rates for Black people, but makes the robustness rates for White people very high. Second, when there is enough data, targeting $\Delta$ based on domain knowledge is promising for being able to certify a larger fraction of the data, as we see with Georgia. 

\section{Conclusions and broader impacts}\label{sec:conclusions}

We defined the label-bias robustness problem and showed how to certify pointwise bias-robustness for linear models both exactly and approximately. We evaluated our approaches on three commonly-used benchmark datasets, showing that (i) our approaches can certify bias robustness and (ii) there are commonly gaps in bias-robustness, particularly for smaller demographic groups. These results merit further study of bias-robustness, and suggest that our technique may be valuable for ML practitioners wishing to explore limitations of their models. 

\paragraph{Limitations} 
A major limitation of our work is that we use the closed-form least-squares formula, and not a commonly used surrogate like gradient descent. 
Additionally, if any fairness interventions (e.g., weighting the dataset or forcing behavior such as equal odds) are applied, then our algorithm is not directly applicable, and the trends we observed may not hold. Future work is needed for bias-robustness certification for more complex algorithms.
The restriction of data bias to labels is another limitation, since data may be biased or inaccurate for other reasons - e.g., under-representation of some groups or poorly transcribed features. We used Income and COMPAS as real-world examples where bias may have impacted labeling; however, future work should include domain experts' input and focus on certifying robustness to a wider range of bias types.

\paragraph{Broader impacts} We hope that bias-robustness certification can help ML practitioners assess whether the outputs of their models are reliable. Specifically, failure to achieve certification is a sign that the prediction may have been tainted by biased data and should not be taken at face value. Given infrequent non-robustness it may be appropriate to deploy a model and then abstain or defer to a secondary decision-making process upon encountering a non-robust test point; widespread non-robustness may necessitate a different modeling technique, different training data, or both. Clearly, more work is needed to make this vision a reality, both on the technical side (adapting our approach for commonly used implementations) and on a sociotechnical scale (deciding how to proceed when decisions are not robust). Additionally, more work is needed to understand what realistic bias models look like in different domains, which is a question that must be addressed by careful thought and reflection of data collectors and domain experts.

The certification technique that we propose could be leveraged adversarially to construct minimal label-flipping data poisoning attacks, a possibility we explore in the appendix. However, we feel that these attacks would not be high-impact compared with other techniques, since they require full dataset access and -- in their most effective form -- target a specific test point. Another risk of our technique is fairwashing. That is, a bias-robustness certification may be taken as evidence that a machine learning prediction is reliable and trustworthy, when in reality, the results are invalid (either due to a poorly specified bias model, or to extraneous factors like a poorly-defined machine learning task). It is important that anyone who uses our certification approach continues to think critically about other potential failure points.




\renewcommand*{\bibfont}{\small}
\printbibliography


\newpage
\appendix

\section{Additional details from sections \ref{sec:3}-\ref{sec:approx}}

\paragraph{Exact approach in the binary case}
As discussed in \cref{sec:exact}, the exact approach is sound and precise in the binary 0/1 case. Even though we technically allow perturbed labels to take on any real value in [0,1], in practice, all labels will remain either 0 or 1 due to the linearity of maximal perturbations. 

\paragraph{Computing $\Theta$}
We walk through an example of computing $\Theta$ directly. Recall that $\matrprodc=(\x^\top \x)^{-1}\x^\top$. 

\begin{example}\label{ex:approx1}
Suppose $\y=(1,-1,2)$, $\Delta=[-1,1]^3$, and $\l=2$. Given $\matrprodc=\begin{pmatrix}1 & 2 & 1 \\ -1 & 0 &2 \\ 2 &1 & 0 \end{pmatrix}$, we have 
$$ \thetaset =  \matrprodc\y \cup 
\left\{\matrprodc\begin{pmatrix}a \\ b \\2 \end{pmatrix} \right\} \cup
\left\{\matrprodc\begin{pmatrix}a \\ -1 \\c \end{pmatrix}\right\} \cup
\left\{ \matrprodc\begin{pmatrix}1 \\ b \\c  \end{pmatrix}\right\} $$

for $a\in[0,2]$, $b\in[-2,0]$, and $c\in[1,3]$. 

Note that $(3,6,3)^\top\in\thetaset$ and $(4,5,2)^\top\in\thetaset$, but their midpoint $(3.5, 5.5, 3.5)^\top\notin\thetaset$, thus, $\thetaset$ is non-convex.
\end{example}

\section{Complete algorithms}

\cref{alg:exact-full} is the complete algorithm for the approach described in \cref{sec:exact}. Note that this algorithm supersedes \cref{alg:exact}. 

We define the \emph{negative potential impact} $\rho^-$ as the minimization counterpart of $\rho^+$, i.e., $\rho^-$ is the maximal negative change that perturbing $\yp_i$ can have on $\matrprod\y$.

\begin{algorithm}[ht]
    \caption{Solve \protect{\cref{eq:max}} by finding a perturbation of $\y$ that maximally increases $\matrprod\y$ and then a perturbation of $\y$ that maximally decreases $\matrprod\y$}
    \begin{algorithmic}
        \Require $\matrprod\in\mathbb{R}^n, \y\in\mathbb{R}^n, \Delta\in\intdomain^n$ with $0\in\Delta$, $\l\geq 0$
        \State $\yl\gets\y$ and $\yu\gets\y$
        \If{$z_i\geq 0$}
            \State $\rho^+_i \gets \matrprod_i\delta_i^u\ $, $\ \ \rho^-_i \gets \matrprod_i\delta_i^l$
        \Else 
            \State $\rho^+_i \gets \matrprod_i\delta_i^l\ $,  $\ \ \rho^-_i\gets \matrprod_i\delta_i^l$
        \EndIf
        \State Let $\rho^+_{i_1},\ldots,\rho^+_{i_l}$ be the $\l$ largest elements of $\rho^+$ by absolute value
        \ForEach{$\rho^+_{i_j}$}
            \If{$z_{i_j}\geq 0$}
                \State $(\yuscal)_{i_j} \gets (\yuscal)_{i_j} + \delta_{i_j}^u$
            \Else
                \State $(\yuscal)_{i_j}\gets (\yuscal)_{i_j} + \delta_{i_j}^l$
            \EndIf
        \EndFor
        \State Let $\rho^-_{i_1},\ldots,\rho^-_{i_l}$ be the $\l$ largest elements of $\rho^-$ by absolute value
        \ForEach{$\rho^-_{i_j}$}
            \If{$z_{i_j}\geq 0$}
                \State $(\ylscal)_{i_j}\gets (\ylscal)_{i_j} + \delta_{i_j}^l $
            \Else
                \State $(\ylscal)_{i_j} \gets (\ylscal)_{i_j} + \delta_{i_j}^u$
            \EndIf
        \EndFor
        \State $V=[\matrprod\yl, \matrprod\yu]$
    \end{algorithmic}\label{alg:exact-full}
\end{algorithm}

We also present the complete algorithm for the approximate approach in \cref{alg:approx}. 

\begin{algorithm}
    
    \caption{Computing $\thetaset^a$}
    \begin{algorithmic}
        \Require $\matrprodc\in\mathbb{R}^{m\times n}, \y\in\mathbb{R}^n, \Delta\in\intdomain^n$ with $0\in\Delta$, $\l\geq 0$
        \State $\ \thetaset^a\gets[0,0]^m$
        \For{$i$ \textrm{ in range} $m$}:
            \State $\yl\gets\y$, $\ \ \yu\gets\y$
            \If{$c_{ij} < 0$}
                \State  $\rho^+_j \gets \matrprodc_{ij}\delta_j^l\ $,  $\ \ \rho^-_j\gets \matrprodc_{ij}\delta_j^u$
            \Else
                \State $\rho^+_j\gets\matrprodc_{ij}\delta_j^u\ $,  $\ \ \rho^-_j\gets \matrprodc_{ij}\delta_j^l$
            \EndIf
            \State Let $\rho^+_{k_1},\ldots,\rho^+_{k_l}$ be the $\l$ largest elements of $\rho^+$ by absolute value.
            \ForEach{$\rho^+_{k_j}$}
                \If{$c_{i{k_j}}\geq 0$}
                    \State $(\yuscal)_{k_j} \gets (\yuscal)_{k_j} + \delta_{k_j}^u$
                \Else
                    \State $(\yuscal)_{k_j} \gets (\yuscal)_{k_j} + \delta_{k_j}^l$
                \EndIf
            \EndFor
            \State Let $\rho^-_{k_1},\ldots,\rho^-_{k_l}$ be the $\l$ largest elements of $\rho^-$ by absolute value.
            \ForEach{$\rho^-_{k_j}$}
                \If{$c_{i{k_j}} \geq 0$}
                    \State $(\ylscal)_{k_j} \gets (\ylscal)_{k_j} + \delta_{k_j}^l$
                \Else
                    \State $(\ylscal)_{k_j} \gets (\ylscal)_{k_j} + \delta_{k_j}^u$
                \EndIf
            \EndFor
        
            \State $\thetaset^a_i \gets [\mathbf{c}_i\yl, \mathbf{c}_i \yu]$
        \EndFor
    \end{algorithmic}\label{alg:approx}

\end{algorithm}
 
\section{Omitted proofs}

Proof of \cref{lemma:1}
\begin{proof}
(Contradiction) Suppose $\matrprod\yp\notin[\hat{y}-\rr, \hat{y}+\rr]$. Then, without loss of generality, there is some $\yp\in\Bias_{\l,\Delta}$ such that $\matrprod\yp>\hat{y}+\rr$. Thus, we can modify the prediction of $\mathbf{x}$ by more than $\rr$ given $\Delta$-allowable label perturbations, so $\mathbf{x}$ is not $(\rr,\Bias_{\l,\Delta})$-robust.
\end{proof}

Proof of \cref{lemma:2}
\begin{proof}
Let $V=[\min_{\yp\in\Bias_{\l,\Delta}(\y)}\matrprod\yp, \max_{\yp\in\Bias_{\l,\Delta}(\y)}\matrprod\yp]$. Then if $V^u>\matrprod\y+\rr$, by \cref{lemma:1}, $\mathbf{x}$ is not $(\rr,\Bias_{\l,\Delta}$-robust. A similar argument holds for $V^l$. 

Suppose $\mathbf{x}$ is $(\rr,\Bias_{\l,\Delta}$-robust. Then there is no $\yp\in\Bias_{\l,\Delta}(\y)$ with $\matrprod\yp>\matrprod\y+\rr$. Thus $\max_{\yp\in\Bias_{\l,\Delta}(\y)}\matrprod\yp= V^u\leq \matrprod\y + \rr$. A similar argument holds for $\matrprod\y-\rr$ and $V^l$.
\end{proof}

Proof of \cref{thm:exact}
\begin{proof}
Our goal is to show that $V^u$, the output of \cref{alg:exact}, satisfies $V^u=\max_{\yp\in\Bias_{\l,\Delta}(\y)}\matrprod\yp$. 

Suppose that there is a label perturbation that we did not make that would have perturbed $\matrprod\yu$ more. That is, for some label perturbation $y_{i_j}^u=y_{i_j}^u+\delta_{i_j}^u$, there was a different label $y_m$, with $m\notin\{i_1,\cdots,i_{\l}\}$ with $\matrprod_m\delta_m^u>\matrprod_{i_j}\delta_{i_j}^u$. But then $m$ would have been in $\{i_1,\cdots,i_{\l}\}$. So this is not possible, thus, $\yu$ satisfies $\argmax_{\yp\in\Bias_{\l,\Delta}(\y)} \matrprod\yp$, and $V^u=\matrprod\yu$.
\end{proof}

Proof of \cref{thm:ratio}
\begin{proof}
Suppose $\mathbf{x}$ is $(\rr, \Bias_{\l,\Delta})-$robust. Consider $\y'\in\Bias_{\l,\Delta}(\y)$, supposing that labels $y'_{i_1}, \ldots,y'_{i_l}$ were perturbed by $d_{i_1},\ldots,d_{i_l}$ from $\y$. 
Note that $d_{i_j}\in\delta_{i_j}$, and that $\matrprod\y\in[\matrprod\y-\rr, \matrprod\y+\rr]$. 
But we can write $\matrprod\y'= \matrprod\y + \matrprodscal_{i_1}y'_{i_1} + \cdots + \matrprodscal_{i_l}y'_{i_l}$. Thus, $\matrprodscal_{i_1}y'_{i_1} + \cdots + \matrprodscal_{i_l}y'_{i_l} \in [-\epsilon,\epsilon]$. 
If we instead create $\y''$ by perturbing the labels $i_1,\ldots,i_l$ of $\y$ by $cd_{i_1}, \ldots, cd_{i_l}$ to satisfy $\y''\in\Bias_{\l,c\Delta}(\y)$, we will have $\matrprod\y''\in[\matrprod\y - c\rr, \matrprod\y+c\rr]$.  
\end{proof}

Proof of \cref{thm:approx}
\begin{proof}
(i) Consider $\theta^*\in \thetaset$. Then, $\theta^*=\matrprodc\y^*$ for some $\y^*\in\Bias(\y)$. Consider coordinate $i$ of $\theta^*$. We have $\theta^*_i\in\thetaset^a$ because the lower bound of $\theta^a$ was chosen to be $\min_{\y'\in\Bias(\y)}\matrprodc_i\y'$. Similarly, the upper bound of $\theta^*$ was chosen to be $\max_{\y'\in\Bias\y}(\matrprodc_i\y')$. Therefore, $\theta^*_i=\matrprodc_i\y^*\in\thetaset^a_i$.

(ii) Suppose there is another box ${{\thetaset}^a}'$ with ${{\thetaset}^a}'_i\subset {\thetaset}^a_i$. 
WLOG, assume that the upper bound of ${\thetaset^a}'_i$ is strictly less than the upper bound of ${\thetaset}^a_i$. 
But ${\thetaset}^a_i=\max_{\y'\in\Bias(\y)}(\matrprod_i\y')$, which means that $\y^*=\max_{\y'\in\Bias(\y)}\matrprodc \y'$ has $\matrprodc_i\y^*$ greater than the upper bound of ${\thetaset^a}'_i$, and thus ${\thetaset^a}'$ is not a sound enclosure of $\thetaset$.
\end{proof}
\section{Additional experimental results}

\subsection{More details on the robustness-accuracy tradeoff}
\cref{fig:exact_app_compare} shows the robustness for MNIST 1/7, Income, and COMPAS at four different accuracy levels. For the most-accurate version (column 1), we see that the over-approximate certifiable robustness rates are very low, especially for MNIST 1/7, where we cannot certify anything. As the accuracy level drops, the amount we can certify grows for both the exact and approximate approaches, but this trend is more pronounced for the approximate certification.
\Cref{tab:all_data_tolerances} shows the same information numerically.

\input{graphs/exactapproxall}

\begin{table}[t]
\small
\centering
\caption{Robustness rates for different numbers of label-flips for various classification datasets. Each entry is the percentage of test-set elements that are certifiably robust against any training-data set perturbation involving up to a the specified percentage of labels.}
\label{tab:all_data_tolerances}
\begin{tabular}{l|l|rrrrrrrrrr}\toprule
\multirow{2}{*}{Dataset} & Accuracy & \multicolumn{10}{c}{Bias level as a percentage of training dataset size}  \\
                        & \multicolumn{1}{c|}{loss} & 0.25  & 0.5 & 1.0  & 2.0  & 3.0  & 4.0 & 5.0 & 6.0 & 7.0 & 8.0 \\\midrule
\multirow{6}{*}{MNIST 1/7}  &   0 & 96.3 & 93.1 & 84.4 & 60.8 & 38.8 & 23.3 & 13.1 &  7.0 &  2.4 & 0.4 \\
                            & 0.2 & 96.7 & 93.6 & 85.8 & 63.2 & 42.5 & 27.0 & 16.1 &  9.7 & 4.6 & 1.1 \\
                            & 0.5 & 97.7 & 95.3 & 90.8 & 79.1 & 64.1 & 49.1 & 37.2 & 27.1 & 21.0 & 15.5 \\
                            & 1.0 & 98.5 & 97.3 & 94.6 & 89.8 & 84.8 & 78.0 & 69.5 & 60.8 & 51.7 & 44.9 \\
                            & 1.5 & 99.2 & 98.1 & 96.2 & 93.6 & 90.7 & 88.2 & 84.7 & 80.2 & 76.1 & 71.1 \\
                            & 2.0 & 99.2 & 98.6 & 96.9 & 94.1 & 92.0 & 89.4 & 87.0 & 83.1 & 79.6 & 76.2 \\\midrule
\multirow{6}{*}{Income}     &   0 & 81.4 & 67.8 & 50.7 & 23.3 & 12.1 &  4.8 &  1.7 & 0.7 & 0 & 0  \\
                            & 0.2 & 83.5 & 71.4 & 54.7 & 27.6 & 12.9 &  6.2 &  2.0 &   1.3 &  0& 0 \\
                            & 0.5 & 87.6 & 75.6 & 59.3 & 36.0 & 17.5 &  9.2 &  3.6 &  1.6 &  1.1 & 0 \\
                            & 1.0 & 94.1 & 86.7 & 70.6 & 51.4 & 35.9 & 24.0 & 16.9 &  9.5 &  6.0 &  2.9 \\
                            & 1.5 & 94.1 & 89.3 & 78.4 & 57.6 & 43.2 & 32.5 & 24.0 & 16.4 & 10.9 &  7.3 \\
                            & 2.0 & 94.2 & 89.6 & 81.0 & 58.9 & 45.7 & 34.4 & 25.7 & 18.5 & 11.7 &  9.0 \\\midrule
\multirow{6}{*}{COMPAS}     &   0 & 91.8 & 85.2 & 72.8 & 46.6 & 31.9 & 17.9 &  6.7 &  0.4 & 0 & 0 \\
                            & 0.2 & 92.1 & 85.4 & 73.6 & 46.9 & 32.5 & 18.5 &  7.3 &  0.3 & 0 & 0 \\
                            & 0.5 & 93.7 & 86.2 & 74.3 & 51.7 & 37.8 & 26.0 & 14.4 &  4.2 & 0 & 0 \\
                            & 1.0 & 94.7 & 88.5 & 74.3 & 56.7 & 42.5 & 27.1 & 19.1 & 10.8 & 0.8 & 0 \\
                            & 1.5 & 95.7 & 89.8 & 73.1 & 58.1 & 43.9 & 30.5 & 20.7 & 13.4 & 3.4 &. 0 \\
                            & 2.0 & 94.5 & 89.4 & 76.4 & 61.1 & 48.3 & 37.0 & 30.3 & 19.5 & 13.5 & 3.7 \\

\bottomrule   
\end{tabular}
\end{table}

\subsection{Description of synthetic data}
We created datasets of size 200, 500, or 1000, with 3, 4 or 5 normally distributed features. Feature 1 was $\mathcal{N}(0.5,1)$ for class 1 and $\mathcal{N}(-0.5,1)$ for class 2. Feature 2 was $\mathcal{N}(1,1)$ for both classes. Feature 3 was $\mathcal{N}(-0.5,1)$ and $\mathcal{N}(0.5,1)$ for classes 0 and 1, respectively, feature 4 was $\mathcal{N}(1,1)$ and $\mathcal{N}(-1,1)$ for the two classes, and feature 5 was $\mathcal{N}(0,1)$ for both classes. 

\paragraph{Demographics data} We used 4 different distributions to generate data in $\mathbb{R}^2$ for the majority group with label 1, on the minority group with label 1, in the majority group with label 0, and the minority group with label 0. Details of distribution means and covariances are available in our code repository.


\subsection{Synthetic dataset results}\label{sec:synth-experiments}

Compared with the real-world datasets, the synthetic datasets could withstand much higher rates of label-flipping bias while maintaining robustness. We hypothesize that this is because the synthetic datasets are much `easier' in general, as they lack real-world noise and obey a linearly-separable data distribution, which is not always true of real-world data.

We performed two types of experiments with synthetic data. The first probes how the number of features and dataset size impact robustness, while the second investigates how robust different simulated demographic groups are relative to subgroup representation. 
These results validate our experimental approach because the results are very intuitive. Namely, there was higher robustness when the datasets were larger and when there were fewer features, as is true in ML more broadly.

We first discuss the impact of general data properties, and then turn our attention to the demographics-based experiments.

\subsubsection{Dataset properties}
We found that the dataset size and the number of features both impact label-bias robustness.

\paragraph{Dataset size}
When there are just three features, the dataset size does not significantly impact robustness. However, the addition of a fourth, and especially a fifth, feature leads to a positive correlation between the size of the dataset and label-bias robustness, as summarized in \cref{tab:synth_dataset_size}. 
We hypothesize that the size of the dataset matters less given fewer features because it takes less data to be confident in the predictor and to attain a certain level of stability with fewer features, that is, fewer potential confounders. 

\paragraph{Number of features}
Adding additional features reduces robustness, as shown in \cref{tab:synth_dataset_size}. 

\begin{table}[t]
\small
\centering
\caption{Robustness rates for synthetic datasets of different sizes and number of features.}
\label{tab:synth_dataset_size}
\begin{tabular}{ll|rrrrr}\toprule
\multirow{2}{*}{Dataset size} & \multirow{2}{*}{Number of features} & \multicolumn{5}{c}{Bias level as a \% of training dataset size}  \\
                        & & 4  & 8  & 12  & 16  & 20 \\\midrule
200 & 3 & 83.8 &75.5  & 69.8 & 66.8 & 65.0  \\
500 & 3 &  83.7 & 75.2 & 70.3 & 68.0 & 67.0 \\
1000 & 3&  84.0 & 75.5   & 69.6 & 68.1 & 67.0  \\
200 & 4 & 81.5 & 71.3 & 64.0 & 61.3 & 61.0  \\
500 & 4 & 81.7 & 70.5   & 65.3 & 63.3 & 61.6  \\
1000 & 4 & 82.9 & 71.9   & 66.1  & 63.3 & 62.6  \\
200 & 5 &  78.0  &  66.5 & 59.8 &56.8  & 56.0  \\
500 & 5 & 79.5 & 69.1   & 60.6 & 57.2 & 56.8 \\
1000 & 5 & 79.9 & 67.6 & 62.2 & 59.5 & 58.5  \\
\bottomrule   
\end{tabular}
\end{table}

\paragraph{Results for approximate robustness}
Approximate robustness behaves similarly upon perturbing the dataset size and is very sensitive to the number of features. We can certify higher label-bias robustness when there are fewer features, and this trend holds not only in absolute terms but also relative to exact certification. For example, given 10\% label bias, when there are three features, we can certify 79.1\% of the samples as robust using the exact procedure and 60.2\% as robust using the approximate version -- an 18.9\% gap. For four and five features, we can certify 76.4\% and 74.7\%, respectively, of samples exactly, and 45.5\% and 31.7\% using the approximate technique, leading to certification gaps of 30.9\% for four features and 43\% for five features.

This trend can also be directly inferred from the algorithm: because we bound each coordinate of $\thetaset$ (i.e., the coefficient for each coordinate) individually, a larger number of features will inherently lead to less robust bounds.


\subsubsection{Demographic experiments}\label{sec:demo}
We found that subgroup representation (i.e., the size of the demographic minority) was very influential for label-bias robustness. 

\paragraph{Subgroup representation}
As shown in \cref{fig:demo1}, when the minority demographic group forms a smaller percent of the total population, that demographic group is less robust. By contrast, the majority group sees only  slightly higher robustness rates when its majority is larger.

\input{graphs/demo_synth}


Additionally, as shown in \cref{fig:demo2}, we found that robustness rates for the minority group are much more variable across folds when the minority group is smaller. We hypothesize that this trend is related to the size of the dataset; that is, when the minority group is very small, we suspect that there is just not enough data to make generalizable conclusions.

\subsection{Dataset variation experiments}

\paragraph{Altering the dataset size}
We repeated the Income and COMPAS experiments on random subsets of 20-80\% of the original data. For Income, the robustness rates are indistinguishable across all dataset sizes. However, for COMPAS, larger dataset sizes yield higher robustness, e.g., given 0.5\% label bias, we can certify 81.5\% of test points as robust on the full dataset but only 49.8\% on the 20\%-size dataset.
We suspect that Income is similarly robust at different sizes because it is much larger: even the smallest subset, containing 20\% of the full dataset, is larger than all of COMPAS.

\paragraph{COMPAS demographics}
The version of the COMPAS dataset that we use has the racial breakdown of 34.2\% White, 51.4\% Black, 8.2\% Hispanic, and 6.2\% other. Analysis of our results from \cref{sec:exactresults} show that the robustness rates between White and Black defendants diverge significantly for moderate (>1.5\%) label-bias, as shown in \cref{tab:compasdemo}.

\begin{table}[t]
\small
\centering
\caption{Robustness rates by race, COMPAS.}
\label{tab:compasdemo}
\begin{tabular}{l|rrrrrrrrrrr}\toprule
\multirow{2}{*}{} & \multicolumn{11}{c}{Bias level as a percentage of training dataset size}  \\
                         & 0.1  & 0.25  & 0.5  & 0.75  & 1.0  & 1.5  & 2.0  & 3.0  & 4.0 & 5.0 & 6.0 \\\midrule
Black               & 96.9 & 92.7 & 86.5 & 81.2 & 71.9 & 49.5 & 40.3 & 28.9 & 11.8 & 1.3 & 0.7 \\
White                   & 93.7 & 88.9 & 83.1 & 74.8& 73.0 & 66.9 & 56.6 & 37.0 & 30.4 & 17.7 & 0.0  \\
\bottomrule   
\end{tabular}
\end{table}

We tried balancing the dataset demographically by selecting a subset of the data that was 50\% White people and 50\% Black people. This re-balancing increases the robustness of the previously less-represented group (i.e., White people) while having no impact on the average robustness for Black people. Clearly, demographic representation is not sufficient for equal robustness rates. We leave a further exploration of what properties of population subgroups impact robustness for future work, and note that domain knowledge is likely to be important in making these discoveries.

\paragraph{Income demographics} \Cref{tab:statesdemo} shows the demographic make-up of various states' data from the Income dataset.

\begin{table}[t]
\small
\centering
\caption{Summary of data download by state from the Folktables Income task.}
\label{tab:statesdemo}
\begin{tabular}{l|rrr}\toprule
State & training $n$ & \% White & \% Black \\ \midrule
Georgia & 40731 & 67.6 & 23.9 \\
Louisiana & 16533 & 70.9 & 23.5 \\
Maryland & 26433 & 63.6 & 23.5 \\
Oregon & 17537 & 86.4 & 1.4 \\
Wisconsin & 26153 & 92.7 & 2.6 \\
\bottomrule   
\end{tabular}
\end{table}

\subsection{Running time}
\cref{tab:timedata} shows the running time of our techniques, as evaluated on a 2020 MacBook Pro with 16GB memory and 8 cores. These times should be interpreted as upper bounds; in practice, both approaches are amenable to parallelization, which would yield faster performance. 

\begin{table}[t]
\small
\centering
\caption{Running time for exact and approximate experiments. The exact experiments flip labels for each data point until the sample is no longer robust. The approximate experiments flip 200 labels (which is enough to bring the robustness to 0\%).}
\begin{tabular}{l|rrrrrr}\toprule
\multirow{2}{*}{Dataset} & \multicolumn{2}{c}{\num{100} samples} & \multicolumn{2}{c}{\num{1000} samples} & \multicolumn{2}{c}{\num{10000} samples} \\
     & Exact & Approx. & Exact & Approx. & Exact & Approx. \\\midrule
Income    & 5.25 & 4.79 & 44.90 & 6.33  & 442.47 & 22.02 \\
Compas    & 1.07 & 2.32 &  6.49 & 2.44  &  51.04 & 10.90 \\
MNIST 1/7 & 4.05 & 3.83 & 24.78 & 5.99 &  448.50 & 24.33 \\\bottomrule                          
\end{tabular}\label{tab:timedata}
\end{table}

\subsection{Connection to attacks}\label{sec:attacks}
As linear regression is often not used in practice for classification datasets, we were interested in seeing whether our approach could identify effective label-bias attacks that generalize to other machine learning algorithms, e.g., logistic regression. 

We targeted 100 samples in each Income and MNIST 1/7. We performed six different attacks: one was the minimal attack needed to change the sample's binary prediction, and the others flipped a fixed percentage (1\%, 2\%, 3\%, 4\% or 5\%) of the test samples. 

The results are summarized in \cref{tab:attacks}. Notably, if we target MNIST 1/7 test samples and are allowed to flip at least 4\% of the labels, the attack will carry over to logistic regression over 90\% of the time. The attacks were not as effective on Income. 

\begin{table}[thbp]
\small
\centering
\caption{Each entry shows the percentage of test samples for which the attack generated by our method, flipping the specified amount of training labels, changed the classification. For example, the minimal perturbation needed to change a sample's classification under our approach resulted in a classification change for out-of-the-box logistic regression 4\% of the time. }
\begin{tabular}{l|rrrrrr}\toprule
          & minimal & 1\% & 2\% & 3\% & 4\% & 5\%  \\\midrule
Income    & 4       & 3   & 5   & 5   & 5   & 5             \\
MNIST 1/7 & 18      & 36  & 71  & 84  & 94  & 98     \\\bottomrule        
\end{tabular}\label{tab:attacks}
\end{table}

\end{document}